%% file: main.tex
\documentclass{article}
\usepackage{amssymb}

\usepackage{graphicx}
\usepackage{subfigure}
\usepackage{amsmath}
\usepackage{mathtools}
\usepackage{amsthm}
\usepackage{microtype}
\usepackage{booktabs}
\usepackage{hyperref}
\usepackage{color}
\usepackage{algorithmic}
\usepackage[table]{xcolor}
\usepackage[ruled,vlined]{algorithm2e}
\usepackage[capitalize,noabbrev]{cleveref}
\usepackage[textsize=tiny]{todonotes}

\usepackage[preprint]{corl_2025} 

\title{Decision MetaMamba: Enhancing Selective SSM in Offline RL with Heterogeneous Sequence Mixing}

\author{
  Wall Kim\\
  Samsung Electronics\\
  Hwaseong, South Korea\\
  \texttt{wall.kim@samsung.com} \\
  \And
  Chaeyoung Song\\
  Seoul National University of Science and Technology\\
  Seoul, South Korea\\
  \texttt{cy.song@seoultech.ac.kr}\\
  \And
  Hanul Kim\\
  Seoul National University of Science and Technology\\
  Seoul, South Korea\\
  \texttt{hukim@seoultech.ac.kr}\\
}

\begin{document}
\maketitle

\input{sections/0_abstract}


\input{sections/1_intro}

\input{sections/2_related}

\input{sections/3_method}

\input{sections/4_experiments}

\input{sections/5_conclusion}

\clearpage
\bibliography{main}
\input{sections/X_appendix}

\end{document}

%% file: sections/0_abstract.tex
\begin{abstract}
Mamba-based models have drawn much attention in offline RL. However, their selective mechanism often detrimental when key steps in RL sequences are omitted. To address these issues, we propose a simple yet effective structure, called Decision MetaMamba (DMM), which replaces Mamba’s token mixer with a dense layer-based sequence mixer and modifies positional structure to preserve local information. By performing sequence mixing that considers all channels simultaneously before Mamba, DMM prevents information loss due to selective scanning and residual gating. Extensive experiments demonstrate that our DMM delivers the state-of-the-art performance across diverse RL tasks. Furthermore, DMM achieves these results with a compact parameter footprint, demonstrating strong potential for real-world applications. Code is available at \url{https://github.com/too-z/decision-metamamba}
\end{abstract}

%% file: sections/1_intro.tex
\section{Introduction} \label{sec:intro}


Offline Reinforcement Learning (RL) can be framed as a sequence modeling problem, where the goal is to predict actions based on pre-collected trajectories of states, actions, and rewards without interacting with the environment in real time\citep{levine2020offlinereinforcementlearningtutorial}. A prominent example of this approach is the Decision Transformer (DT)\citep{chen2021decision}, which leverages the Transformer architecture and introduces hindsight matching\citep{furuta2021generalized} by replacing rewards with return-to-go ($rtg$) as input. Building on this formulation, there has been growing interest in exploring a wide range of modern sequence modeling architectures, including both Transformers and State-Space Models. Among State-Space Models, Mamba\citep{gu2023mamba} has emerged as a particularly promising alternative, demonstrating superior performance and efficiency in language modeling\citep{park2024can}, while also generalizing well across other domains such as vision\citep{zhu2024vision, liu2024vmambavisualstatespace}, graphs\citep{wang2024graph}, and time series\citep{schiff2024caduceus, Wang2024mamba, patro2024simba}. Given its strong modeling capacity and computational efficiency in long-range sequence tasks, Mamba presents a compelling candidate for advancing offline RL beyond Transformer-based approaches. 

However, although effective in various sequence modeling tasks, both Transformers and Mamba may suffer from information loss due to their reliance on selective modeling, which emphasizes specific steps in the sequence\citep{gulati2020conformer}. Furthermore, while self-attention\citep{vaswani2017attention} excels at capturing long-range dependencies, it is less effective in modeling the local transition dynamics characteristic of Markov processes, where proximate steps exert greater influence\citep{resnick2013adventures}. These limitations are particularly pronounced in sparse reward settings, where the limited inductive bias provided by the $rtg$ necessitates a greater dependence on the transition model to infer optimal actions.


We propose Decision MetaMamba (DMM), a heterogeneous sequence mixing model that combines a dense layer-based local mixer with Mamba in a complementary design. DMM consists of two components: a Dense Sequence Mixer (DSM) for capturing local dependencies, and our modified Mamba for modeling interactions across the entire sequence. The DSM functions as the local mixer by flattening and concatenating input embeddings within a local window, and applying an affine transformation to model short-range dependencies. This allows the model to simultaneously consider all input channels and effectively learn short-range patterns. On the other hand, the modified Mamba serves as the global mixer, combining inputs causally and selectively while preserving input shape to capture long-range dependencies. By combining the outputs of both mixers via a residual connection, DMM effectively integrates local and global context, mitigating the issue of step omission during inference. We evaluate our DMM on various benchmarks in offlineRL, including MuJoCo~\citep{wawrzynski2009cat, erez2012infinite}, AntMaze, and Franka Kitchen~\citep{gupta2019relay} datasets from D4RL~\citep{fu2020d4rl}. Despite its simplicity, DMM achieves superior performance over recent state-of-the-art methods, while maintaining high efficiency suitable for resource-constrained edge devices and robotic platforms.


We summarize our main contributions as follows:
\begin{itemize}    
    \item We design a Dense Sequence Mixer (DSM) to perform local mixing via dense affine transformation over flattened input windows, effectively modeling short-range transition dynamics.
    \vspace{0.1cm}
    \item We introduce a Decision MetaMamba (DMM) that integrates the DSM and a modified Mamba to capture local and global dependencies and to preserve causal modeling and long-range interactions in offline RL settings.
    \vspace{0.1cm}
    \item We evaluate DMM on multiple offline RL benchmarks, including MuJoCo, AntMaze, and Franka Kitchen from D4RL. DMM consistently outperforms recent Transformer- and SSM-based methods, while using significantly fewer parameters due to its efficient structure. 
\end{itemize}

\vspace{0.2cm}

\begin{figure}[!t]
\centering
    \subfigure{
        \includegraphics[height=4cm]{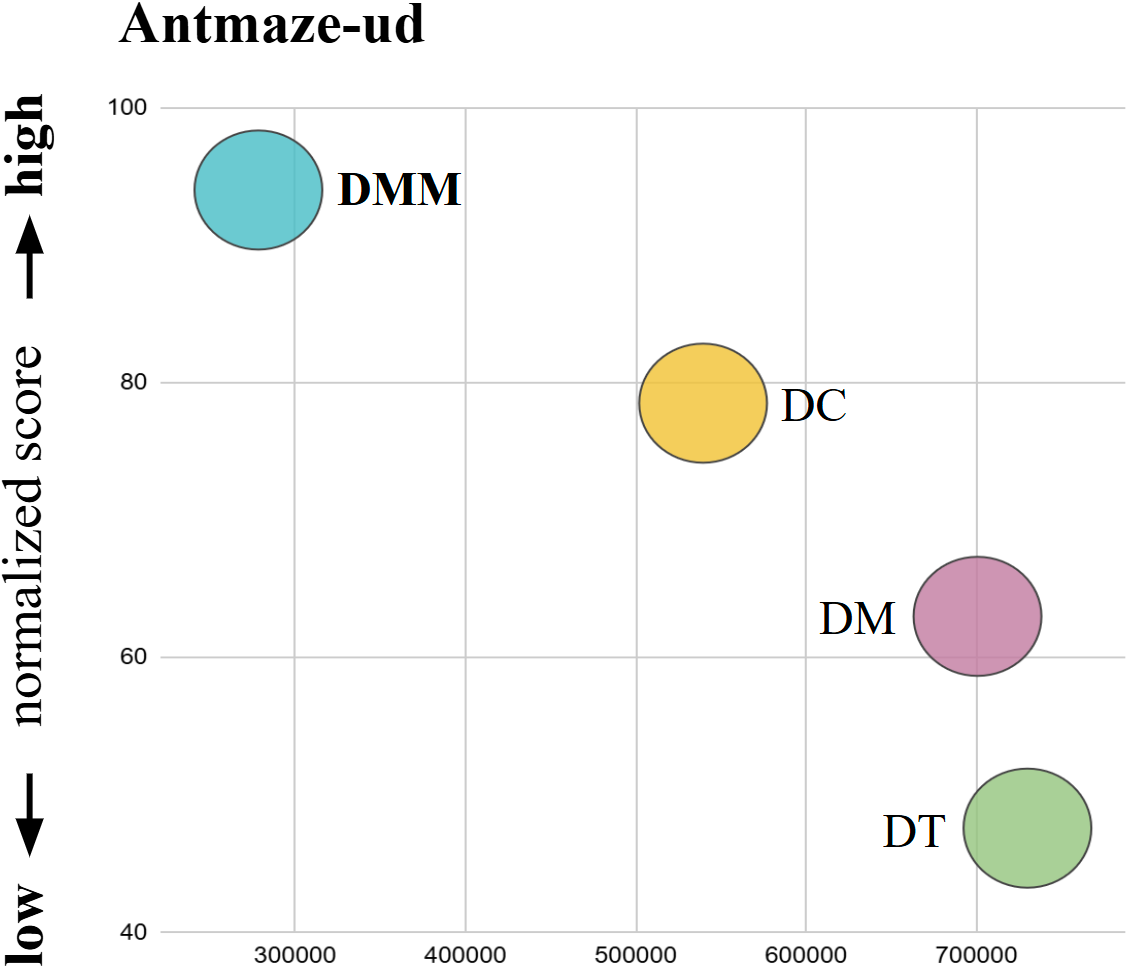}
        \label{fig:l1wodsm}
    }
    \hspace{2cm}
    \subfigure{
        \includegraphics[height=4cm]{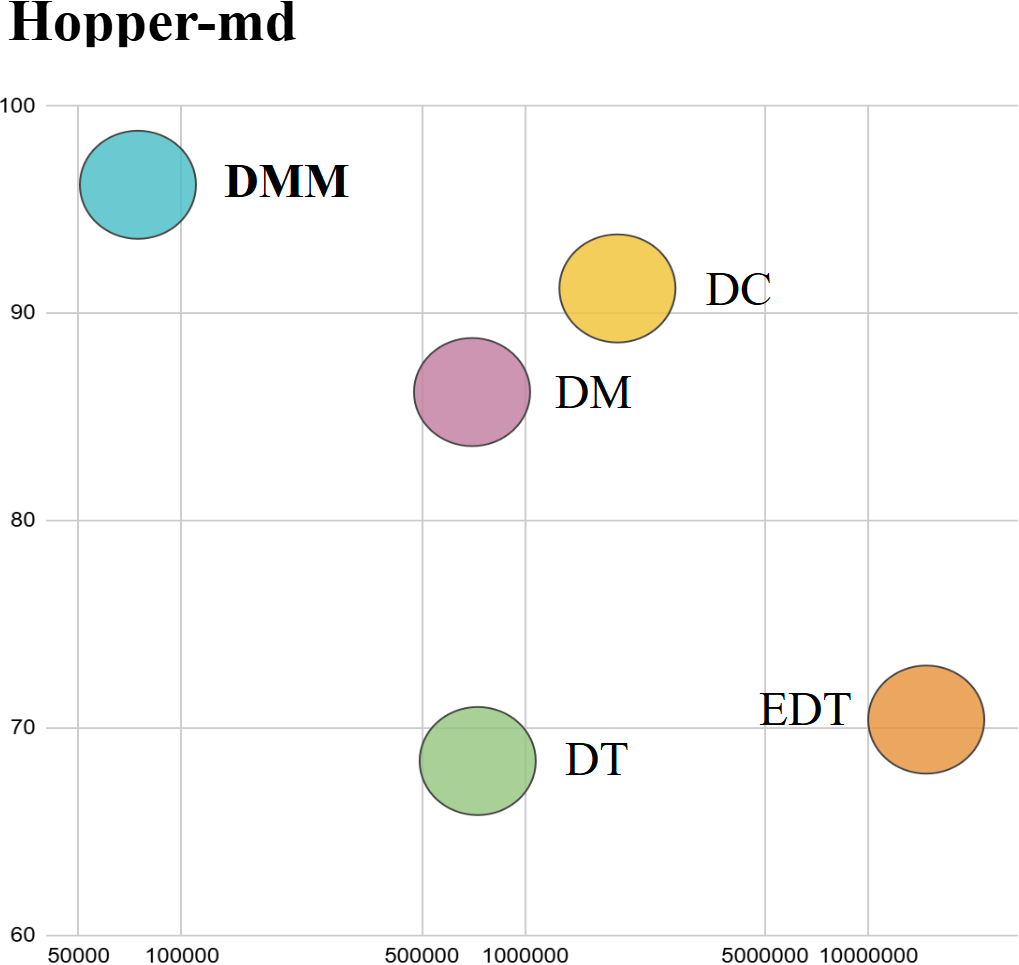}
        \label{fig:l1withdsm}
    }
\caption{Comparison of model size and performance with recent transformer- and SSM-based methods: DC~\citep{kim2024decision}, DT~\citep{chen2021decision}, EDT~\citep{wu2024elastic} and DM~\citep{ota2024decision}. For hopper-md (medium), the x-axis is logarithmic; others use a linear scale.}
\label{fig:performance_n_parameters}
\end{figure}


%% file: sections/2_related.tex
\section{Related Work} \label{sec:related}

\paragraph{Offline RL Decision Models}

Offline RL methods fall into three main categories. Value-based methods like Batch-Constrained Q-learning \cite{fujimoto2019off}, Conservative Q-learning \cite{kumar2020conservative}, and Implicit Q-learning \cite{kostrikov2022offline} constrain policies to the behavior policy. Value-free methods, such as Multi-Game Decision Transformer \cite{lee2022multi} and Trajectory Transformer \cite{janner2021offline}, directly learn policies to mitigate out-of-distribution issues. Model-based methods generate synthetic data using learned dynamics, e.g., Koopman Q-learning \cite{weissenbacher2022koopmanqlearningofflinereinforcement}. 

Transformer-based models like Decision Transformer (DT) \cite{chen2021decision} condition action generation on return-to-go ($rtg$), but face trajectory stitching challenges. Extensions like Elastic DT \cite{wu2024elastic} and Critic-Guided DT \cite{wang2024critic} add auxiliary networks to improve performance through trajectory stitching. Decision Convformer \cite{kim2024decision} removes attention via token mixing, achieving strong results without auxiliaries. Diffusion models \citep{ho2020denoising, chi2023diffusion} offer multimodal behavior modeling and stable training, but are compute-intensive. Sequence modeling alternatives include Decision S4 \cite{david2023decision} and Decision Mamba \cite{ota2024decision}, replacing DT’s Transformer with SSMs. Recent advances exploit Mamba’s linear time complexity for sub-goal generation \cite{huang2024decision} or multi-scale processing \cite{lv2024decision}. StARFormer \cite{shang2022starformer} addresses limitations of language models in offline RL by combining local and global sequence modeling: short-term sequences are handled via self-attention, while long-term dependencies are captured with a second Transformer reflecting the Markov property. RvS \cite{emmons2021rvs} demonstrated the effectiveness of MLPs in sequence modeling and highlighted the importance of inductive bias through a simple two-layer feed-forward architecture.

\paragraph{Information loss in Mamba and Transformer}


The Mamba architecture, using residual multiplication and a sigmoid activation, can be viewed as a Gated CNN \citep{dauphin2017language} enhanced with an SSM \citep{yu2024mambaout}. Prior work \citep{bachlechner2021rezero, guo2022beyond} has shown that gating can intensify information loss, as step components with near-zero weighted inputs from selective SSMs remain suppressed. As a result, replacing the Transformer with Mamba alone \citep{ota2024decision} yielded limited offline RL gains.
Efforts across domains have aimed to mitigate information loss in Transformers. For instance, \citet{beltagy2020longformer} preserved local information in long texts via short input ranges, while \citet{gulati2020conformer} used CNNs in local windows to better exploit local features and counter self-attention inefficiencies.

\begin{figure*}[!t]
    \centering
    \subfigure[Mamba]{
        \includegraphics[height=6cm]{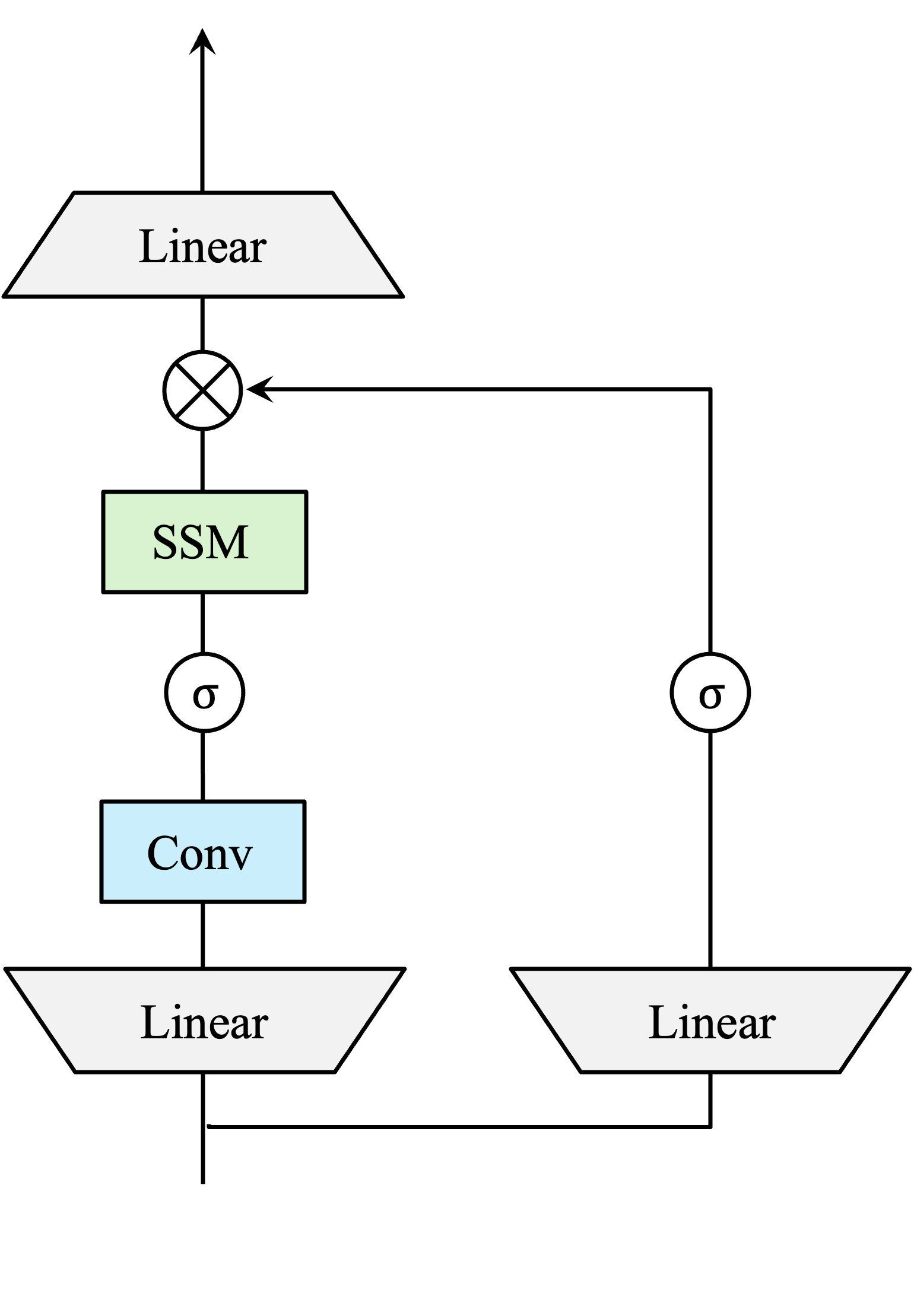}
        \label{fig:arch_mamba}
    }
    \hspace{0.4cm}
    \subfigure[Decision Meta Mamba]{
        \includegraphics[height=6cm]{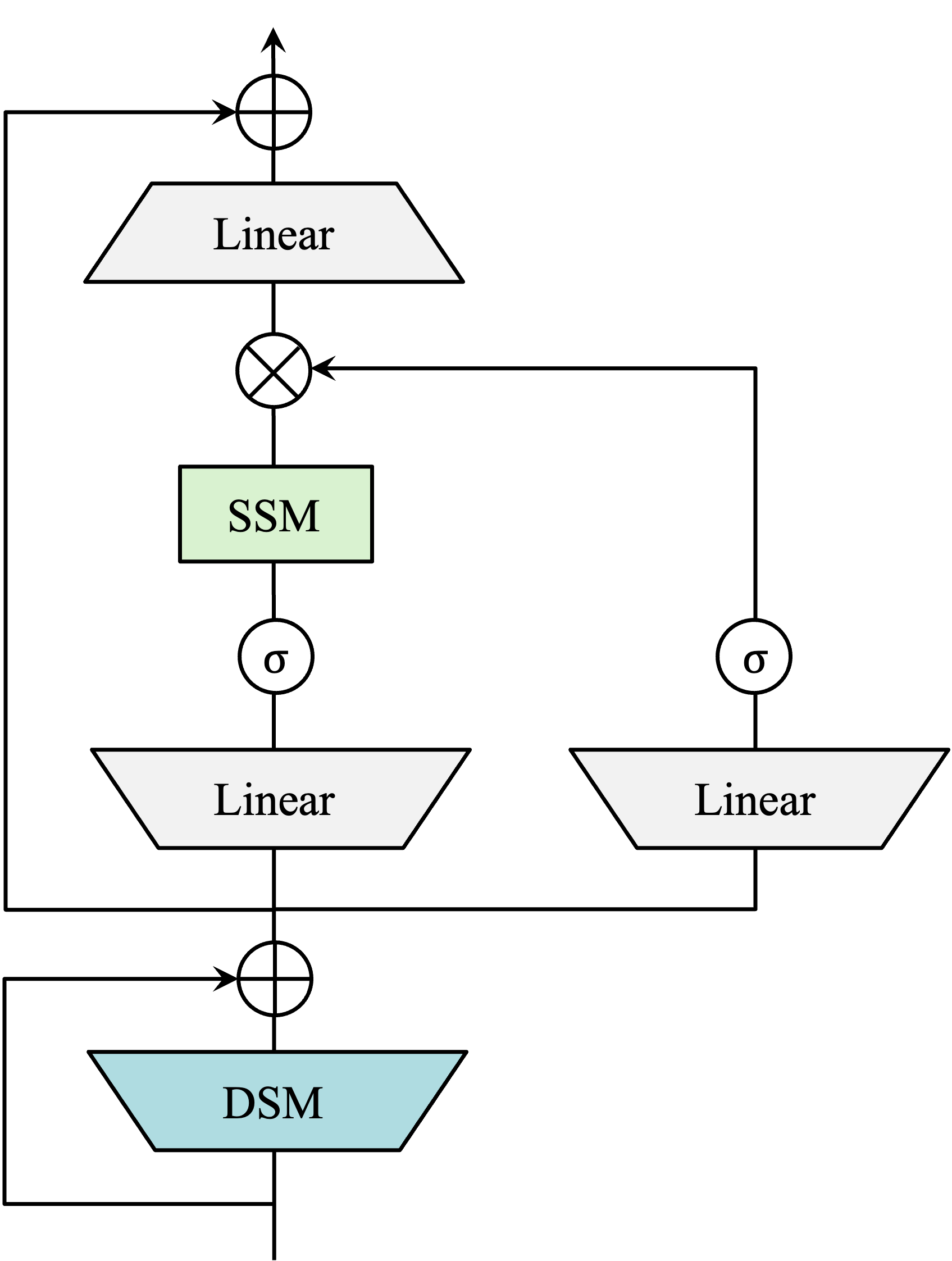}
        \label{fig:arch_dmm}
    }
    \hspace{0.2cm}
    \subfigure[Dense Sequence Mixer]{
        \includegraphics[height=6cm]{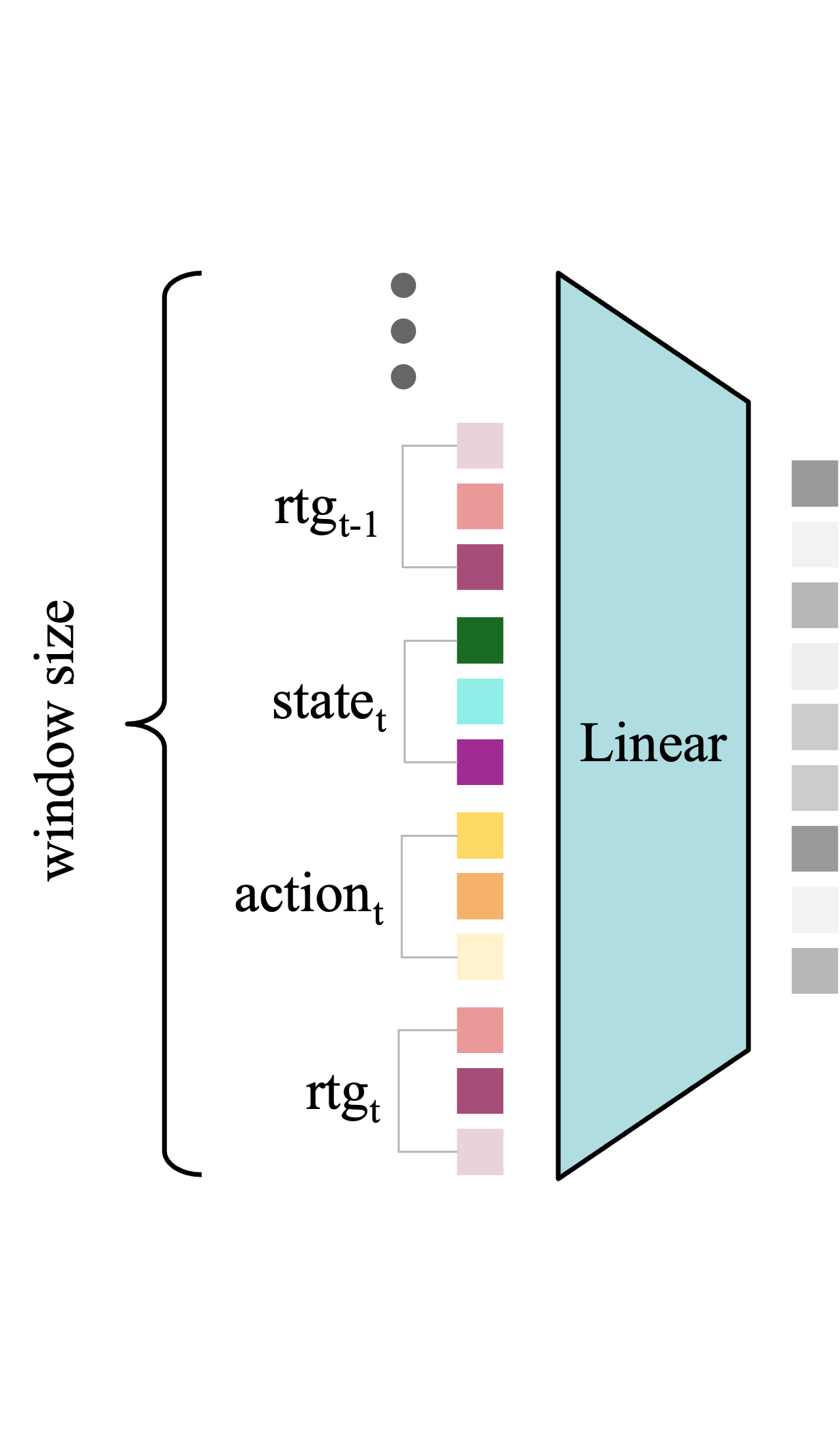}
        \label{fig:arch_mixer}
    }
    \caption{Detailed model structure: (a) Mamba~\cite{gu2023mamba}, (b) Decision Mamba, and (c) Dense Sequence Mixer (DSM).}
    \label{fig:arch}
    \vspace{0.3cm}
\end{figure*}


%% file: sections/3_method.tex
\section{Methodology} \label{sec:method}

\subsection{Preliminaries}  

State Space Models (SSMs) were originally used in control engineering to model the behavior of systems. In deep learning, by parameterizing and learning the functions of an SSM, we can effectively model the transition dynamics of discretized systems. The input $x_t$ is mapped to a latent space and reconstructed as the output $y_t$:  

\begin{equation} \label{eq:discretization}
\bar{A} = f_A(A,\Delta),\quad \bar{B} = f_B(A, B,\Delta)
\end{equation}
\begin{equation} \label{eq:ssm}
h_t = \bar{A} h_{t-1} + \bar{B} x_t, \quad y_t = C h_t + D x_t
\end{equation}

SSMs, represented by linear differential equations, consist of three time-dependent variables: input $x_t \in \mathbb{C}^{m}$, latent state $h_t \in \mathbb{C}^{n}$, and output $y_t \in \mathbb{C}^{p}$, transformed by four learnable matrices: a transition matrix $A \in \mathbb{C}^{n\times n}$, a control matrix
$B \in \mathbb{C}^{n\times k}$, an output matrix
$C \in \mathbb{C}^{m\times n}$, and a command matrix $D \in \mathbb{C}^{m\times k}$. The direct input influence $D x_t$ is typically omitted but we retain it to reflect the skip connection that directly adds $x_t$ to the selective scan SSM output.. For discrete sequential data, SSMs are adapted using zero-order hold (ZOH)~\cite{gu2023mamba} for discretization (\Cref{eq:discretization}), convolutional kernels~\cite{gu2021combiningrecurrentconvolutionalcontinuoustime} for parallelization, and structured transition matrices~\cite{gu2021efficiently} for computational efficiency.

\begin{figure*}[!t]
    \centering
    \subfigure{
        \includegraphics[height=3cm]{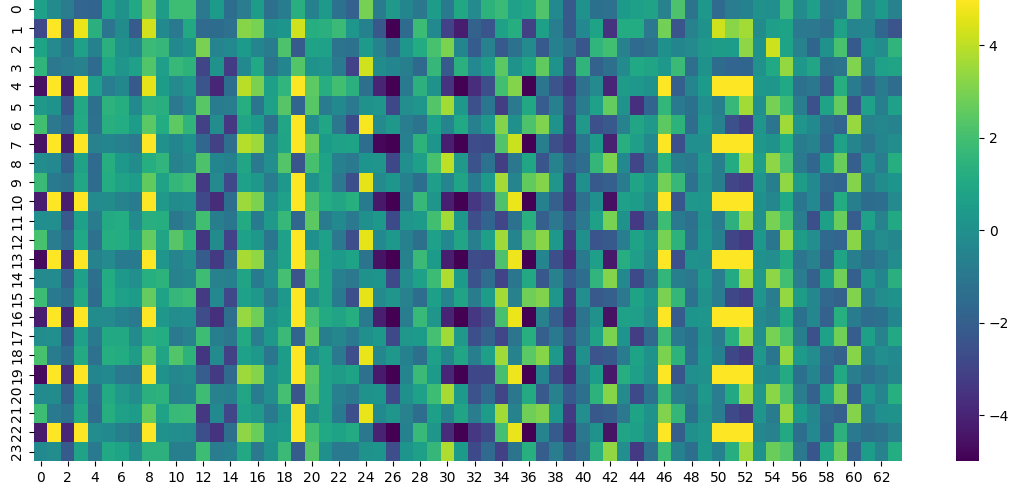}
        \label{fig:l1wodsm}
    }
    \hspace{0.4cm}
    \subfigure{
        \includegraphics[height=3cm]{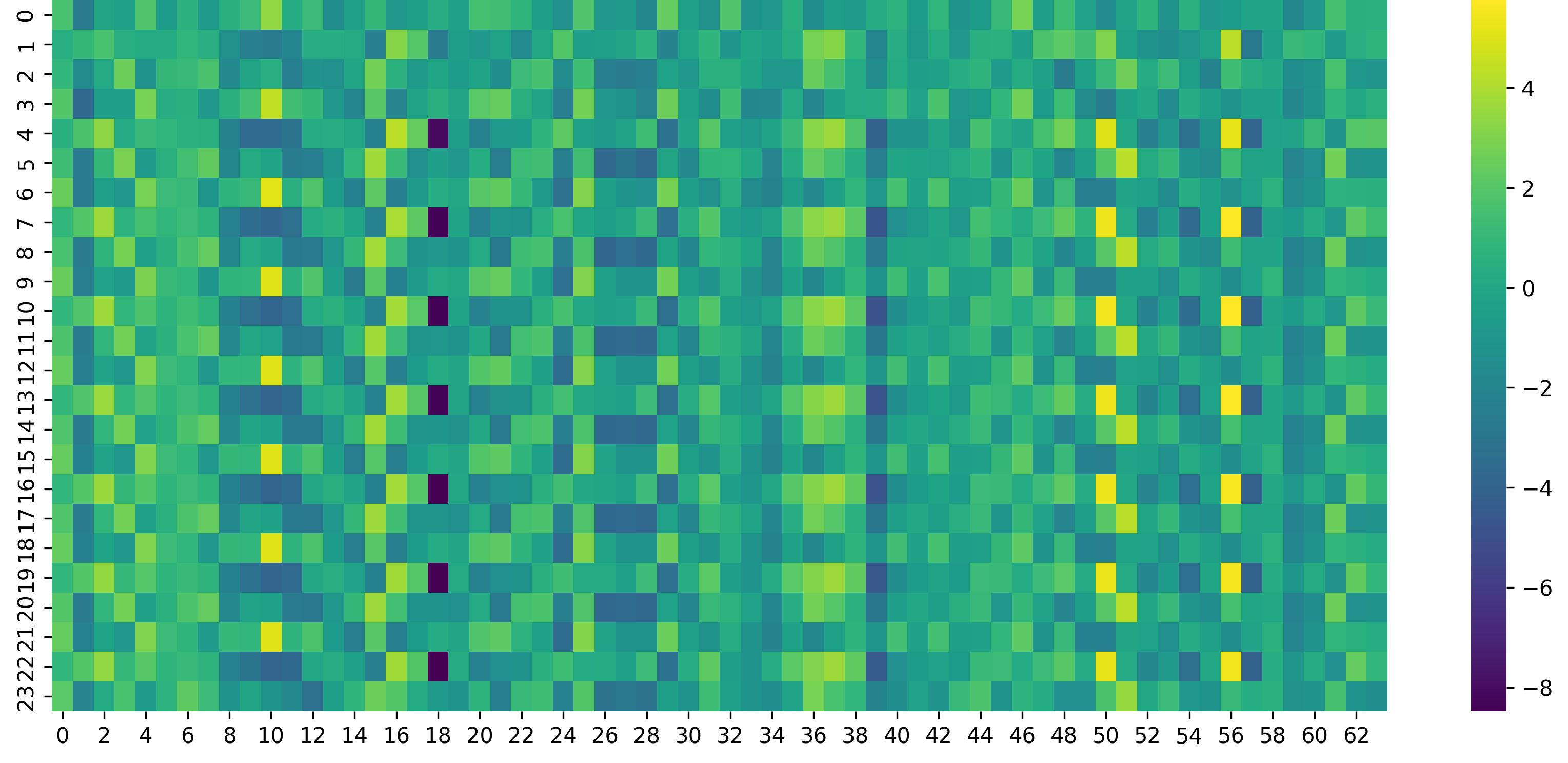}
        \label{fig:l1withdsm}
    }
    \caption{Heatmap visualization of the output tensor activation. The y-axis denotes step components in the repeating order of state, action, and $rtg$, starting from index 0. The x-axis corresponds to the embedding dimension. Left: output from Mamba~\cite{gu2023mamba}. Right: output from DSM + modified Mamba.}
    \label{fig:l1norm}
\end{figure*}

\subsection{Motivations}



Selective scan SSM integrates inputs selectively during training and inference. As shown in~\Cref{fig:l1norm}, Mamba’s activation map highlights amplified action steps. Like self-attention in Transformers, this enhances content awareness by downweighting less informative tokens, similar to how words like "the" are minimized in language models. However, in offline RL, excluding tokens risks losing immediate context. Some state and $rtg$ vectors approach zero in the activation map, diminishing their output influence. Since optimal actions depend on state and $rtg$ dynamics, this can cause critical information loss. In Mamba, this issue stems from residual multiplication, a gating mechanism combining selective SSM outputs with activation-processed inputs, which may erase step information already suppressed by the SSM or activation function~\citep{bachlechner2021rezero}. Performance drops with ReLU~\citep{kim2024decision}, which zeros out negative inputs, further support the presence of information loss(Appendix \ref{app:activation_functions}). We identify this as a structural limitation in Mamba and propose addressing it to enhance RL performance.

\subsection{Decision MetaMamba}
Next, we propose Decision MetaMamba (DMM), motivated by our observations. First, we introduce a new local sequence mixer, termed a dense sequence mixer (DSM). Unlike a selective SSM, which serves as a global sequence mixer to capture long-range token dependencies, DSM explicitly leverages dependencies between contiguous tokens. We then replace the depth-wise convolution-based sequence mixer from the original Mamba with DSM, to develop our DMM block. 

\paragraph{Dense Sequence Mixer}
Let $\mathbf{x}_t = (\mathbf{r}_t, \mathbf{s}_t, \mathbf{a}_t)$ be the input tokens at time step~$t$, where $\mathbf{r}_t$, $\mathbf{s}_t$, and $\mathbf{a}_t$ denote the return-to-go ($rtg$), state, and action tokens, respectively. For each time step~$t$, we define the input sequence $X_t$ as $X_t = (X_{t-k+1}, X_{t-k+2}, \ldots, X_t)$, where $k$ is the window size. \Cref{fig:arch_mixer} illustrates the structure of the proposed DSM. Given the input sequence $X_t$, DSM concatenates all tokens and then performs a linear projection. 

Compared to Mamba’s 1D depth-wise convolution for local sequence mixing, our DSM processes each token in its entirety, rather than handling a single value per channel. Consequently, DSM more effectively captures dependencies among adjacent tokens, critical for leveraging Markov chain properties in offline RL. Although this design can increase computational and parameter overhead as the window size grows, empirical findings indicate that a small window size usually suffices across diverse offline RL tasks. Additional details can be found in \Cref{sec:experiments}.

\paragraph{Decision MetaMamba Block}
As illustrated in \Cref{fig:arch}, our DMM differs from the original Mamba in three ways. First, it replaces the 1D depth-wise convolution layer with the proposed DSM for local sequence mixing. Second, it repositions the local sequence mixer from inside the Mamba block to the front. Third, the tokens locally mixed by the DSM are fed into the Mamba block’s output using a residual connection. Consequently, our DMM block takes the input sequence $X_t$ and generates its output $Y_t$ as follows:
\begin{equation}
\begin{aligned}
\Tilde{X}_t &= \texttt{LN}(X_t) \\
Z_t &= \Tilde{X}_t + \texttt{DSM}(\Tilde{X}_t) \\
\Tilde{Z}_t &= \texttt{LN}(Z_t) \\
Y_t &= \Tilde{Z}_t + \texttt{ModifiedMamba}(\Tilde{Z}_t)
\end{aligned}
\end{equation}
where $\texttt{LN}(\cdot)$ denotes a layer normalization~\citep{ba2016layer} for stable learning, $\texttt{DSM}(\cdot)$ represents the proposed DSM, and $\texttt{ModifiedMamba}(\cdot)$ refers to the Mamba block without 1D depthwise convolution, respectively. Note that placing DSM at the front allows local token relationships to be leveraged before selective scanning and gating. Additionally, the residual connection from the DSM output to the final Mamba block preserves information that might otherwise be lost through these operations. Furthermore, because Mamba inherently encodes positional information through its sequential state-space formulation, our model does not require any additional positional encoding.

%% file: sections/4_experiments.tex
\begin{table*}[!t]\renewcommand{\arraystretch}{1.2}
\caption{Comparison with state-of-the-art methods in dense-reward environments. Abbreviations: md (medium), mr (medium-replay), and me (medium-expert). Results are obtained from experiment with five random seeds; some are cited from~\citep{ma2023rethinking}. Vertical lines separate value-based, Transformer-based, and SSM-based models. The best and second-best scores are highlighted in \textbf{bold} and \underline{underline}, respectively.}
\label{tab:DRE}
\vspace{0.2cm}
\small
\centering
\begin{tabular}{l|c|cc|cccc|ccc}
\toprule
Dataset & Type & TD3+BC & CQL & DT & QLDT & EDT & DC & DS4 & DM & \textbf{DMM} \\
\midrule
hopper&md&59.3&58.5&68.4&66.5&70.4&\underline{91.2}&	54.7&86.2&\textbf{96.2}\\
hopper&mr&60.9&\textbf{95.0}&82.5&52.1&\underline{89.4}&87.8&	49.6&81.7&\textbf{95.0}\\
hopper&me&98.0&105.4&108.7&94.2&104.7&110.1&	\underline{110.8}&\textbf{111.0}&110.2\\
\midrule
walker2d&md&\textbf{83.7}&72.5&	76.6&67.1&75.8&79.6&	78.0&77.6&\underline{83.6}\\
walker2d&mr&\textbf{81.8}&\underline{77.2}&	65.9&58.2&73.3&77.1&	69.0&72.5&74.1\\	
walker2d&me&\underline{110.1}&108.8&109.0&101.7&107.8&109.2&105.7&108.3&\textbf{110.2}\\
\midrule
halfcheetah&md&\textbf{48.3}&\underline{44.0}&42.8&42.3&43.0&	43.0&42.5&42.8&43.0\\
halfcheetah&mr&\underline{44.6}&\textbf{45.5}&39.1&35.6&37.8&	41.1&15.2&39.8&41.1\\
halfcheetah&me&90.7&91.6&85.5&79.0&89.7&	\textbf{93.1}&\underline{92.7}&90.6&92.6\\
\midrule
mujoco&mean&75.3&77.6&75.4&66.3&76.9&\underline{81.4}&68.6&78.9&\textbf{82.9}\\
\midrule
\multicolumn{2}{c|}{Avg. Rank} & 3.78	&4.11	&5.89	&8.44	&5.56	&\underline{2.89}	&6.44	&4.89	&\textbf{2.33}\\
\bottomrule
\end{tabular}
\vspace{0.3cm}
\end{table*}

\section{Experiments} \label{sec:experiments}
In this section, we carry out comprehensive experiments across diverse offline RL domains to validate the effectiveness of the proposed DMM. We first examine its performance in dense reward environments (DRE), where agents receive frequent rewards throughout the task. Next, we extend our evaluation to sparse reward environments (SRE), where rewards are provided only upon task completion, with all intermediate steps yielding zero reward. Finally, we perform extensive ablation study to analyze the contribution of each DMM component.

For all experiments, we adopt the expert normalized return score~\citep{fu2020d4rl} to measure the performance of DMM. More detailed explanation about this metric and implementation are shown in~\Cref{app:exp_setup}.

\subsection{Results on Dense Reward Environment}
\paragraph{Datasets}
For dense reward environments, we utilize the MuJoCo domain from the popular D4RL~\citep{fu2020d4rl} benchmark, which features a continuous action space with dense rewards. Specifically, we evaluate our DMM on the Hopper, Walker2d~\citep{erez2012infinite}, and HalfCheetah~\citep{wawrzynski2009cat} domains, where the task involves controlling a robot using state information, such as the positions and angles of body parts and joints. All three environments provide rewards at each time step based on the agent's actions and the object's state, making them dense reward environments.

\paragraph{Results}
\Cref{tab:DRE} compares the proposed DMM with recent state-of-the-art methods, including value-based models: TD3+BC~\citep{fujimoto2021minimalist} and CQL~\citep{kumar2020conservative}, transformer-based models: DT~\citep{chen2021decision}, QLDT~\citep{yamagata2023q}, EDT~\citep{wu2024elastic}, DC~\citep{kim2024decision}, and selective SSM-based models: DS4~\citep{david2023decision} and DM~\citep{ota2024decision}.

In \Cref{tab:DRE}, we observe that DMM consistently achieves competitive or superior results across most environments. Notably, DMM outperforms all existing methods in the MuJoCo domain. Specifically, in the Hopper environment, DMM surpasses all value-based models and transformer-based variants, achieving the highest scores. Additionally, in Walker2d, DMM demonstrates performance comparable to the best-performing methods. Consequently, DMM attains the best average ranking in the DRE experiments.

\begin{table*}[!t]
\caption{Comparison with state-of-the-art methods in sparse-reward environments. Abbreviations: um (umaze), ud (umaze diverse), cp (complete), pt (partial), and mx (mixed). Results are obtained from experiment with five random seeds; some are reported from~\citep{ma2023rethinking} and~\citep{badrinath2024waypoint}. Vertical lines separate value-based, Transformer-based, and SSM-based models. The best and second-best scores are marked in \textbf{bold} and \underline{underline}, respectively.}
\label{tab:SRE}
\renewcommand{\arraystretch}{1.2}
\vspace{0.2cm}
\centering
\small
\begin{tabular}{lc|ccc|cccc|ccc}
\toprule
Dataset & Type & TD3+BC & CQL & IQL & DT & QLDT & WT & DC & DS4 & DM & \textbf{DMM} \\
\midrule
AntMaze&um&	78.6&74.0&\underline{87.5}&53.6&67.2&64.9& 85.0 & 63.4&68.0&\textbf{91.0}  \\		
AntMaze&ud&	71.4&\underline{84.0}&62.2&42.2&62.2&71.5&78.5 & 64.6&62.0&\textbf{94.0}  \\		
\midrule
AntMaze&mean&75.0&79.0&74.9&47.9&64.7&68.2&\underline{81.8}&64.0&65.0&\textbf{92.5}\\
\midrule
kitchen&cp&25.0&43.8&62.5&46.5&38.8&49.2&\underline{67.1}&36.3&46.7&\textbf{76.0}  \\		
kitchen&pt&38.3&49.8&46.3&31.4&36.9&63.8&\underline{73.6}&52.9&61.7&\textbf{80.5}  \\
kitchen&mx&45.1&51.0&51.0&25.8&17.7&70.9&\underline{71.8}&47.7&59.3&\textbf{83.0}\\
\midrule
kitchen&mean&36.1&48.2&53.3&34.6&30.5&61.3&\underline{70.8}&45.6&55.9&\textbf{79.8}\\
\midrule
\multicolumn{2}{c|}{Avg. Rank} &6.00 & 4.43	&4.57	&9.43	&8.57	&4.85	&\underline{2.29} &7.57	&6.00	&\textbf{1.00} \\
\bottomrule
\end{tabular}
\vspace{0.3cm}
\end{table*}

\subsection{Results on Sparse Reward Environment}
\paragraph{Datasets}
\noindent For sparse reward environments, we utilize the AntMaze~\citep{fu2020d4rl} and Franka Kitchen~\citep{gupta2019relay} datasets from the standard D4RL benchmark. In the AntMaze environment, agents are tasked with navigating from a fixed initial position to a designated goal within a U-shaped maze (UM). In contrast, the AntMaze umaze-diverse environment (UD) allows random starting positions. The Franka Kitchen~\citep{gupta2019relay} dataset consists of four tasks, where rewards are granted only upon task completion. To ensure consistency with the AntMaze dataset, we modify the reward structure of the Kitchen dataset, as detailed in~\Cref{app:kitchen_modification}. The Kitchen dataset is categorized into three subsets: complete, partial, and mixed target datasets. The complete dataset comprises trajectories in which all four tasks are successfully completed sequentially, while the partial and mixed datasets contain trajectories with incomplete or partially completed tasks. We used modified rewards for the Kitchen dataset, and the details can be found in the Appendix \Cref{app:kitchen_modification}).


\paragraph{Results}
\Cref{tab:SRE} reports the performance of DMM and recent state-of-the-arts, including value-based models: TD3+BC~\citep{fujimoto2021minimalist}, CQL~\citep{kumar2020conservative}, and IQL~\citep{kostrikov2022offline}, transformer-based models: DT~\citep{chen2021decision}, QLDT~\citep{yamagata2023q}, ans WT~\citep{badrinath2024waypoint}, and selective SSM-based models: DS4~\citep{david2023decision} and DM~\citep{ota2024decision}.

In \Cref{tab:SRE}, DMM significantly outperforms all existing methods without exception. It surpasses the second-best method by 13.5 in AntMaze and 18.5 in Kitchen. Furthermore, we see that DMM holds the first place in the average ranking on SRE experiments. These results provide strong empirical evidence supporting the effectiveness of the DMM structure for offline RL in SRE.

Note that SRE presents challenges in credit assignment due to delayed rewards, leading to weak inductive bias and making sequence modeling for offline RL more difficult than in DRE. DMM mitigates this by employing a local sequence mixer to integrate information from consecutive steps, adhering to the Markov property where transitions depend on nearby states. Meanwhile, Mamba selectively incorporates past sequence information, enhancing transition dynamics modeling in offline RL environments with Markov properties. As a result, this improved modeling strengthens action inference, which is particularly critical in SRE due to its limited inductive bias.

\subsection{Analysis on the impact of input component.}
We analyze the impact of each input component - state, action, and $rtg$ - by measering their relative contributions within the input sequence. \Cref{fig:l1norm} shows the the L1 norms of the embeddings, where the norms of state and $rtg$ are consistently 2 to 10 times smaller than those of action across all steps and datasets. This indicates that in input-dependent architectures such as Transformer and Mamba, state and $rtg$ contribute less than action.

\begin{figure*}[!t]
    \centering
    \subfigure{
        \includegraphics[height=3cm]{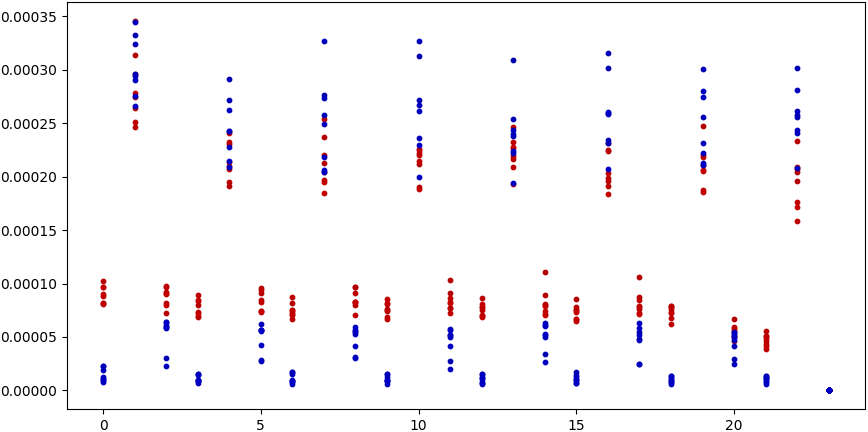}
        \label{fig:gradnorm_sparse}
    }
    \hspace{0.4cm}
    \subfigure{
        \includegraphics[height=3cm]{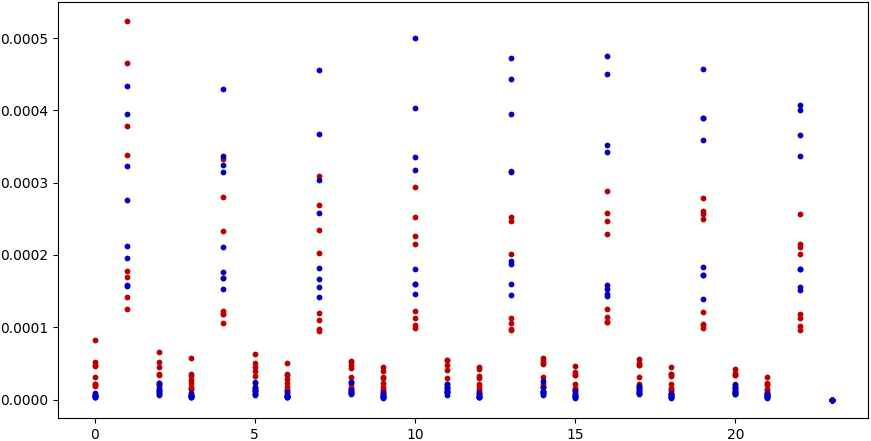}
        \label{fig:gradnorm_dense}
    }
    \caption{Average gradient norms of 24 input features on hopper-md (left) and antmaze-um (right). As the x index increases, the distance from the current step becomes greater, and the sequence is shown in the order of state, action, and $rtg$. Red dots indicate results from DMM, and blue dots from Mamba~\cite{gu2023mamba}. Each dataset includes two batches of 64 samples.}
    \label{fig:gradient_norm}
    \vspace{0.3cm}
\end{figure*}

We then compute gradient norms with respect to each input embedding to assess input influence more directly. As shown in~\Cref{fig:gradient_norm}, in standard Mamba and Transformer, the gradient norms for state and $rtg$ are typically less than one-tenth of that for action. However, with DSM or our modified Mamba, this disparity is reduced, suggesting increased utilization of state and $rtg$. We also observe higher gradients for inputs closer to the current step, aligning with the Markov property. These results suggest that the improved performance of DMM stems from more balanced use of all input components, rather than over-reliance on actions.

\begin{table}[!t]
\renewcommand{\arraystretch}{1.2}
\caption{Comparison of local and global sequence mixers. DMM is the proposed method. Conv replaces the local mixer (DSM) with a 1D depth-wise convolution. Transformer and S4 replace the global mixer (Mamba). Values in parentheses show performance drops relative to DMM.}
\vspace{0.2cm}
\label{tab:analysis_sequence_mixer}
\centering
\small
\begin{tabular}{l|c|c|cc}
\toprule
& DMM & Conv & Transformer & S4 \\
\midrule
Hopper-MD & 96.2 & 94.7(-1.5) & 92.7(-3.5) & 84.6(-11.6)\\
Antmaze-UD & 91.0 & 84.0(-7.0) & 84.0(-10.0) & 81.0(-13.0)\\
Kitchen-MX & 83.0 & 74.3 (-8.7) & 74.5(-8.5) & 77.8(-5.2)\\
\bottomrule
\end{tabular}
\vspace{0.2cm}
\end{table}

\vspace{0.2cm}
\subsection{Analysis on Local and Global Sequence Mixers}
\Cref{tab:analysis_sequence_mixer} presents a comparative analysis of DMM against three baselines to assess the impact of both local and global sequence mixers. First, we replace DSM with a 1D depth-wise convolution from the original Mamba to isolate the contribution of the local mixer. While DSM yields only a marginal improvement in Hopper-MD, it leads to significant performance gains in AntMaze-UM and Kitchen-MX, highlighting its increased effectiveness in SREs that lack rich inductive biases. Second, to evaluate the global mixer design, we replace Mamba with a Transformer and S4. Both alternatives result in consistent performance degradation across all datasets. This demonstrates that Mamba is more effective at capturing long-range dependencies between input steps, making it a superior global sequence mixer for offline RL. Overall, these findings emphasize the importance of connecting local and global sequence mixing through a residual network to maintain and propagate information effectively.

\begin{table}[!t]
\renewcommand{\arraystretch}{1.2}
\caption{Comparison of parameters and cost in methods.}
\label{tab:parameter_efficiency}
\vspace{0.2cm}
\centering
\small
\begin{tabular}{l|c|c|c|c}
\toprule
& \multicolumn{2}{c}{Hopper-MD} & \multicolumn{2}{c}{AntMaze-UM} \\
\midrule
   & param & cost & param & cost \\
\midrule
DMM & 74435 & 773.8 & 21862 & 261.5 \\
\midrule
DT & 727695 & 10638.8 & 730008 & 12166.8 \\
DC & 1858307 & 20376.2 & 539528 & 6347.4 \\
\bottomrule
\end{tabular}
\vspace{0.2cm}
\end{table}

\subsection{Analysis of Parameter Efficiency}
We assess DMM's parameter efficiency by comparing it to recent decision models, Decision Transformer (DT) and Decision Convformer (DC). Specifically, we report each model's parameter count and parameter cost (parameters per score), using hyperparameters from their original papers. \Cref{tab:parameter_efficiency} presents results for Hopper-MD and AntMaze-UM. DMM achieves similar or better performance with far fewer parameters, yielding the lowest parameter cost across both datasets. This efficiency stems from Mamba’s inherently compact architecture and our lightweight offline RL modifications. Mamba also maintains performance with smaller input embeddings. While DC replaces attention with 1D convolutions, it offsets gains by increasing embedding size and MLP channels. Its use of multi-modal token mixing further adds parameters with limited benefit. 

%% file: sections/5_conclusion.tex
\vspace{0.2cm}
\section{Conclusion} \label{sec:conclusion}
In this study, we observed that the Mamba-based decision model exhibited a phenomenon where certain input components were omitted during input-selective sequence reasoning. We demonstrated this experimentally by analyzing differences in gradient norms across step components and attributed the cause to the gating mechanism. To address this, we proposed combining a sequence mixer implemented using a dense layer with a modified Mamba SSM. This design captures local dependencies between adjacent steps and models long-range relationships via the SSM. Our method outperformed state-of-the-art models in most Offline RL tasks and showed notable improvements in sparse reward environments with short context lengths, indicating better modeling of state transitions. Additionally, the DMM achieves strong performance with fewer parameters by using smaller embedding sizes and removing positional encoding, making it well-suited for resource-constrained devices like edge devices and small robots.


\vspace{0.2cm}
\section{Limitation} \label{sec:limit}
While DMM has demonstrated strong performance across most offline RL tasks, further experiments are needed to determine whether additional performance gains can be achieved through online fine-tuning. Moreover, exploring the use of regularization techniques during training may offer further improvements. Lastly, there is a need to investigate methods for enabling constant-time inference. State Space Models (SSMs) like Mamba support constant-time inference due to their sequential computation structure. For DMM to achieve constant-time inference and potentially enhance performance, one approach is to continue appending steps to the end of the sequence during inference. However, this effectively increases the context length over time, resulting in a mismatch with the context used during training. This discrepancy can ultimately lead to performance degradation.

%% file: sections/X_appendix.tex
\newpage
\appendix
\onecolumn

\clearpage
\section{Experiment Details} 
\label{app:exp_setup}

The expert normalized return metric used for model evaluation in the experiment is calculated as follows:
\begin{equation}  
\text{performance} = 100 \times \frac{\text{score}-\text{random score}}{\text{expert score}-\text{random score}}
\end{equation}  
which standardizes the results across datasets, ensuring a normalized range between 0 and 100 for easier comparison. 

In our experiments, the Hopper, Walker2d, and HalfCheetah domains, along with their respective types (MD, MR, ME), as well as the AntMaze and Kitchen domains, including their types (UM, UD, CP, PR, MX), share the following settings. We use the AdamW \cite{loshchilov2018decoupled} optimizer with a learning rate $\eta$ that linearly increases to $1 \times 10^{-4}$ over the first 10,000 steps and then remains constant at $1 \times 10^{-4}$ for the rest of the training, with a weight decay of $1 \times 10^{-4}$. The training process runs for 10 iterations, and each iteration consists of 10,000 optimization steps.

For model training, we use the L2 loss function, following the same approach as the mean squared error (MSE) for continuous actions in \cite{chen2021decision}, and we set the RTG normalization scale to 1,000 to maintain the stability of training. Additionally, the convolution window size used in DSM is set to 6 during training. All experiments were conducted using an Nvidia RTX 3060 12GB GPU on an Ubuntu OS.

For all experiments, we adopt most hyperparameters in decision convformer~\citep{kim2024decision}, to limit the effect of hyper-parameters, except for the dimension of embedding. As confirmed by the experimental results in the Table \ref{tab:half_embedding}, when the input embedding dimension was reduced, the DMM model exhibited robust results without performance degradation, unlike other models. Thus, the adjusted input dimension was set 64 as the default. However, in the sparse reward environments of Antmaze and Franka Kitchen, we employed a 128-dimensional input embedding.

The results of the value-based model in the Mujoco and Antmaze domains are cited from \citep{kim2024decision}, while the Antmaze results for EDT are cited from \citep{ang2024reinformermaxreturnsequencemodeling}. The remaining results are recorded from our own experiments.

\begin{table}[ht]
    \centering
    \small
    \renewcommand{\arraystretch}{1.2}
    \begin{tabular}{ll}
        \noalign{\smallskip}\noalign{\smallskip}\hline
        \textbf{Name} & \textbf{Value} \\
        \hline
        Number of layers  & 3\\
        Batch size  & 64  \\
        Context length $K$ & 8  \\
        Embedding dimension & 64 (32 for Walker2D)\\
        Dropout & 0.1  \\
        Nonlinearity function & SiLU  \\
        Grad norm clip  & 0.25  \\
        Weight decay  & $10^{-4}$  \\
        Learning rate decay  & Linear warmup \\
        Total number of updates & $10^{5}$  \\
        \hline
    \end{tabular}
    \caption{Common hyperparameters of DMM on training of Dense reward environments (Mujoco)}
\end{table}

\begin{table}[ht]
    \centering
    \small
    \renewcommand{\arraystretch}{1.2}
    \begin{tabular}{ll}
        \noalign{\smallskip}\noalign{\smallskip}\hline
        \textbf{Name} & \textbf{Value} \\
        \hline
        Number of layers  & 3\\
        Batch size  & 64  \\
        Context length $K$ & 8  \\
        Embedding dimension & 128\\
        Dropout & 0.1  \\
        Nonlinearity function & SiLU  \\
        Grad norm clip  & 0.25  \\
        Weight decay  & $10^{-4}$  \\
        Learning rate decay  & Linear warmup \\
        Total number of updates & $10^{5}$  \\
        \hline
    \end{tabular}
    \caption{Common hyperparameters of DMM on training of Sparse reward environments (Antmaze, Franka Kitchen)}
\end{table}

Our DMM models are implemented with PyTorch~\citep{paszke2019pytorch} and trained on NVIDIA RTX 3060 GPU. 

\begin{table}[ht]
\small
\centering
\renewcommand{\arraystretch}{1.1}
\resizebox{0.5\textwidth}{!}{
\begin{tabular}{|c|c|c|}
\hline
 & emb dim 128 & emb dim 64\\
\hline
hopper-md & 98.9 & 97.2\\
walker2d-md & 79.7 & 82.1\\
halfcheetah-md & 43.1 & 43.0\\
\hline
\end{tabular}
\hspace{0.2cm}
}
\caption{Performance According to Input Embedding Dimension}
\label{tab:half_embedding}
\end{table}


\begin{figure*}[!t]
    \centering
    \subfigure{
        \includegraphics[height=3cm]{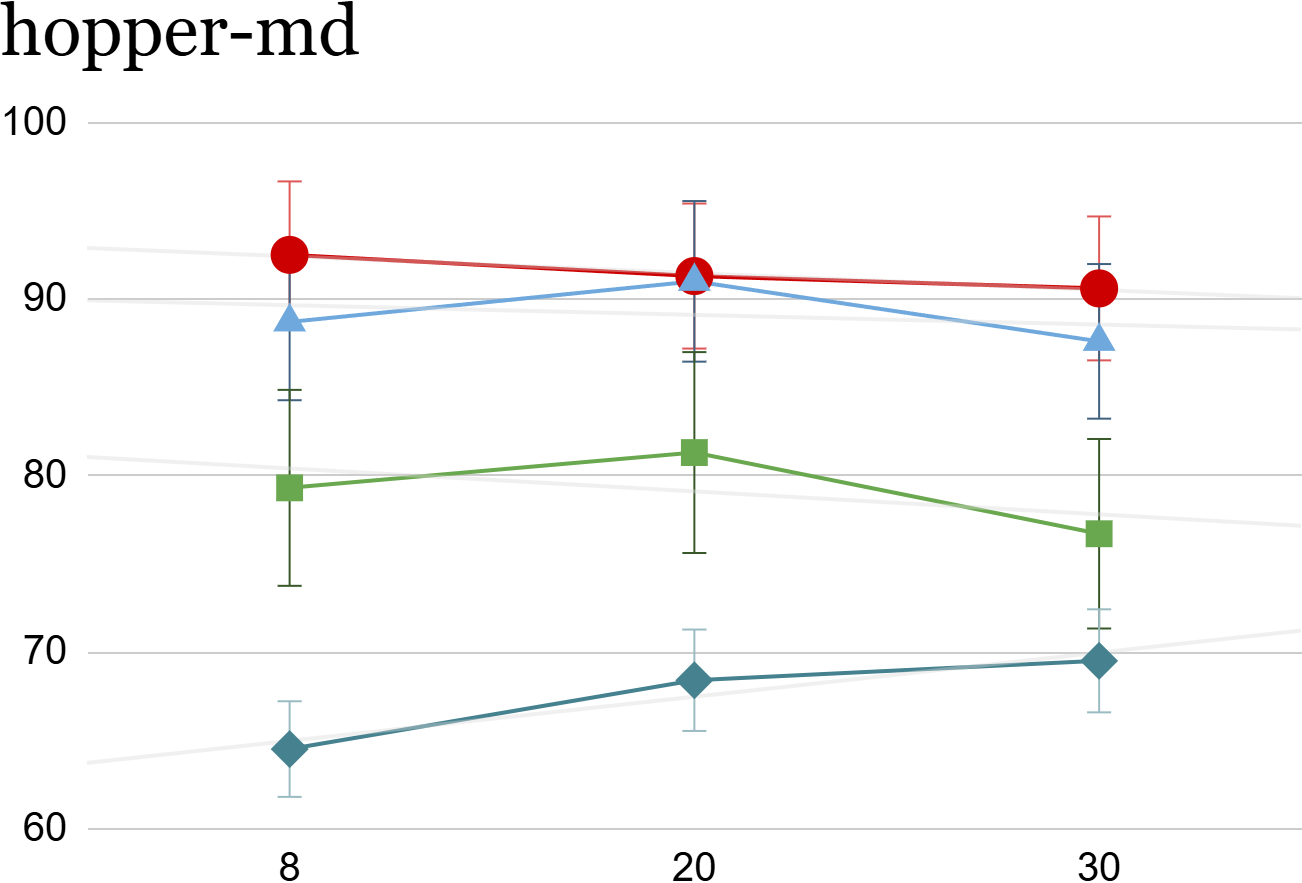}
        \label{fig:context_length1}
    }
    \hspace{-0.2cm}
    \subfigure{
        \includegraphics[height=3cm]{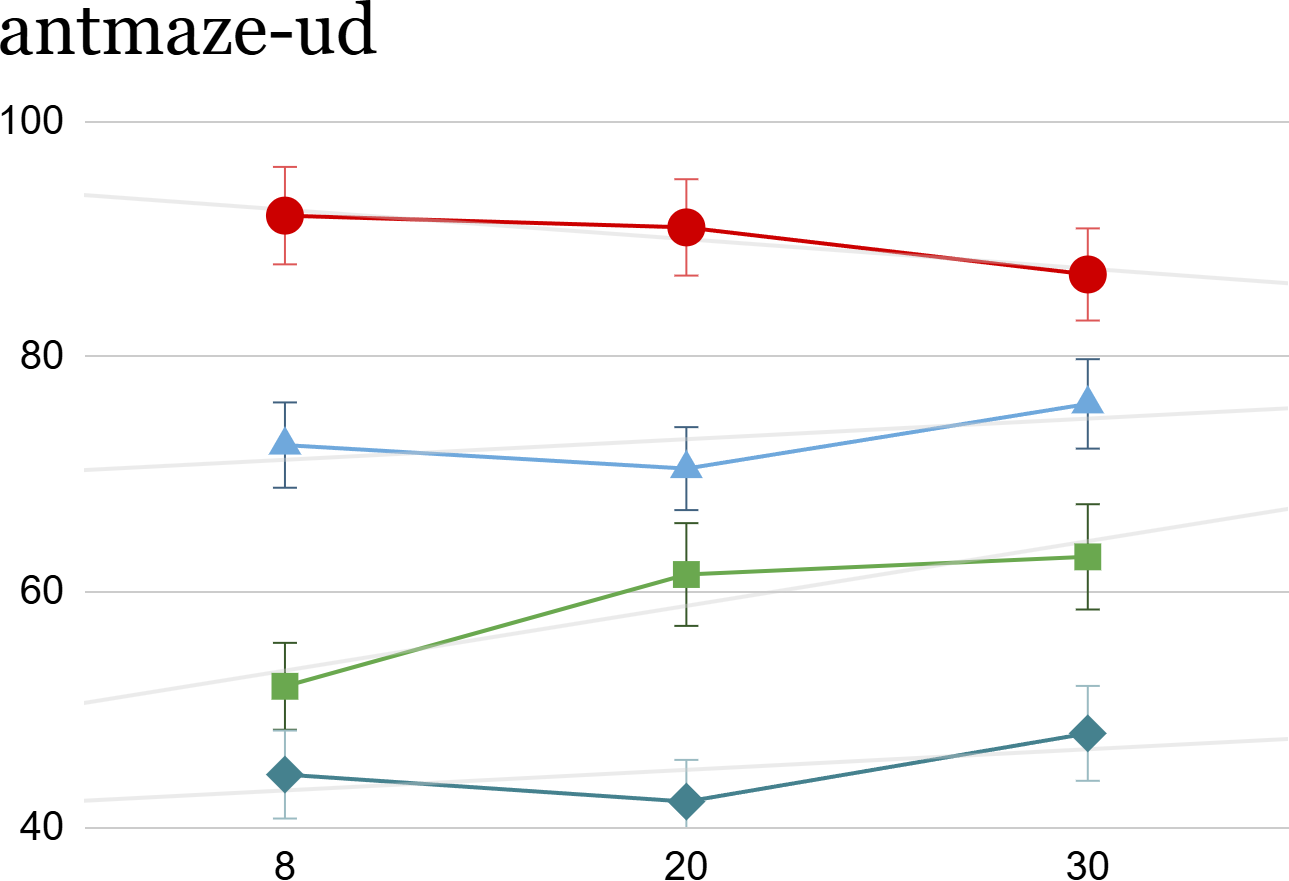}
        \label{fig:context_length2}
    }
    \hspace{-0.2cm}
    \subfigure{
        \includegraphics[height=3cm]{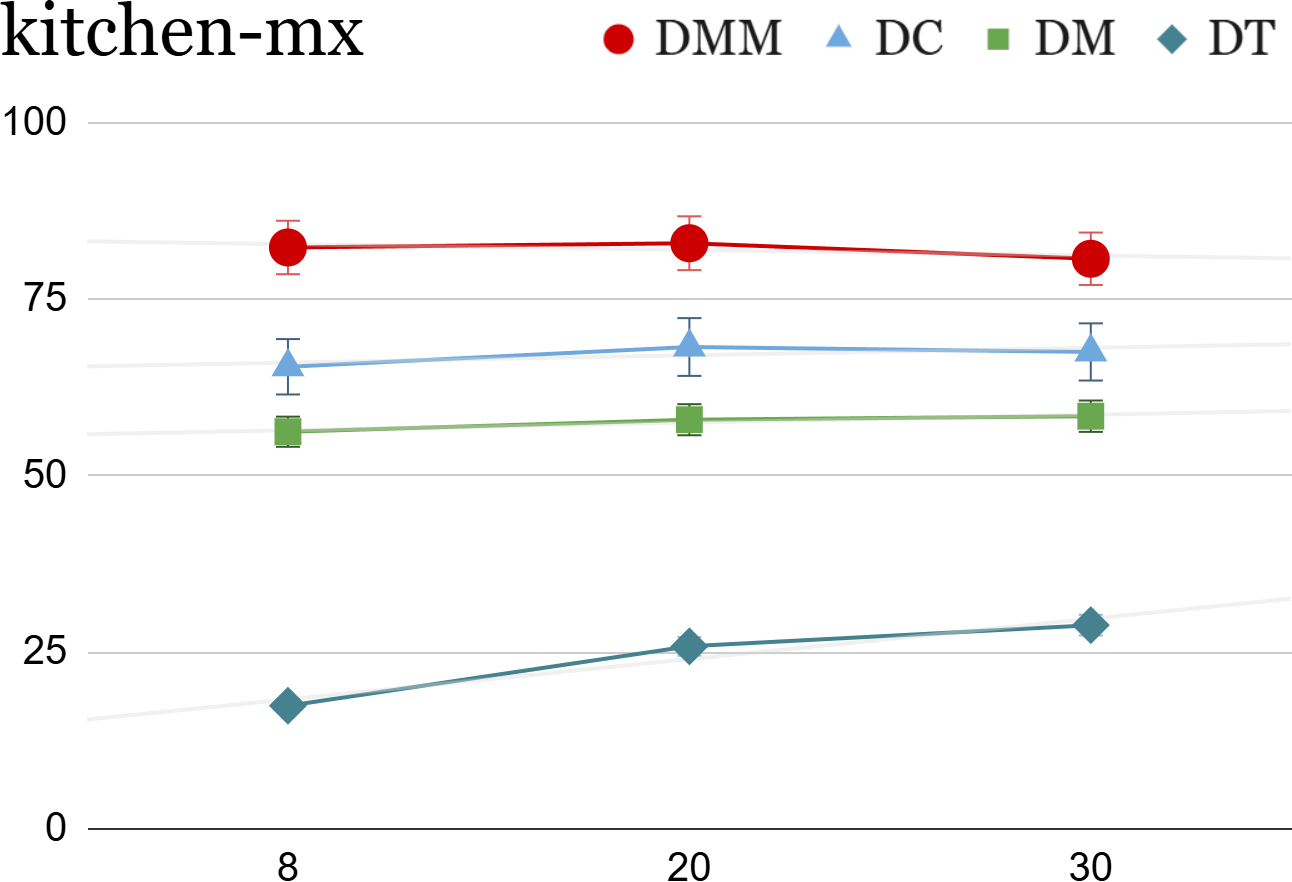}
        \label{fig:context_length3}
    }
    \vspace{-0.1cm}
    \caption{The chart plots evaluation score against context length, with context length on the x-axis and evaluation score on the y-axis. \textit{DMM} achieves the highest performance compared to other models when using a shorter input context length.}
    \label{fig:context_lengths}
\end{figure*}

\section{Analysis on Input Context Length}
\Cref{fig:context_lengths} compares performance across three datasets: Hopper-MD, AntMaze-UD, and Kitchen-MX, based on input context length. Specifically, we evaluate peak performance at context lengths of 8, 20, and 30.  

In~\Cref{fig:context_lengths}, most models acheive peak performance at a context length of 20, while shorter (8) and longer (30) sequences degrade performance, as short contexts hinder action inference and long contexts increase overfitting~\citep{chen2021decision}. In contrast, DMM performs best at a context length of 8, consistently outperforming other models. These results highlight DMM’s ability to effectively model transition dynamics, enabling accurate action inference even with short sequences, which helps mitigate overfitting and improve trajectory stitching in offline RL.

\section{Analysis of Input Context Length and Window Size}
The input sequence length for DSM is defined by the window size, while for Mamba, it is determined by the context length. These parameters control the range of input information processed, and their variation can impact overall performance.

\begin{figure}[!t]
\begin{center}
\centerline{\includegraphics[height=3cm]{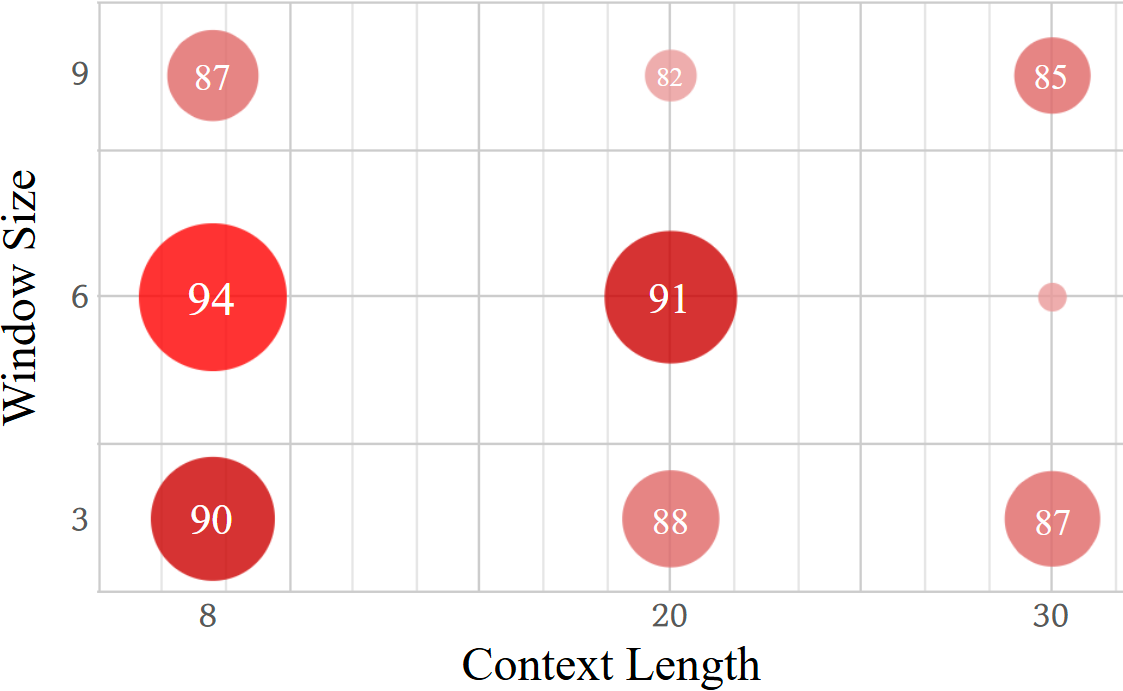}}
\caption{Performance depending on context length and window size. Circle size and number denote peak model performance. }
\label{fig:window_context}
\end{center}
\end{figure}

\Cref{fig:window_context} presents the performance comparison across different window sizes and context lengths in the AntMaze-UD environment. Here, \textit{Context Length} denotes the number of steps, each comprising state, action, and RTG, while \textit{Window Size} specifies the sequence length where state, action, and RTG are treated as a single input. When the window size is 6, DSM processes two steps as input while maintaining causality. This suggests that DSM achieves optimal performance when mixing a sequence approximately 1/4 the length of the entire input.

\section{The Effect of Activation Functions RELU, GELU, and SiLU.} \label{app:activation_functions}
\begin{figure}[ht]
\vskip 0.2in
\begin{center}
\scalebox{0.8}{\includegraphics[width=\columnwidth]{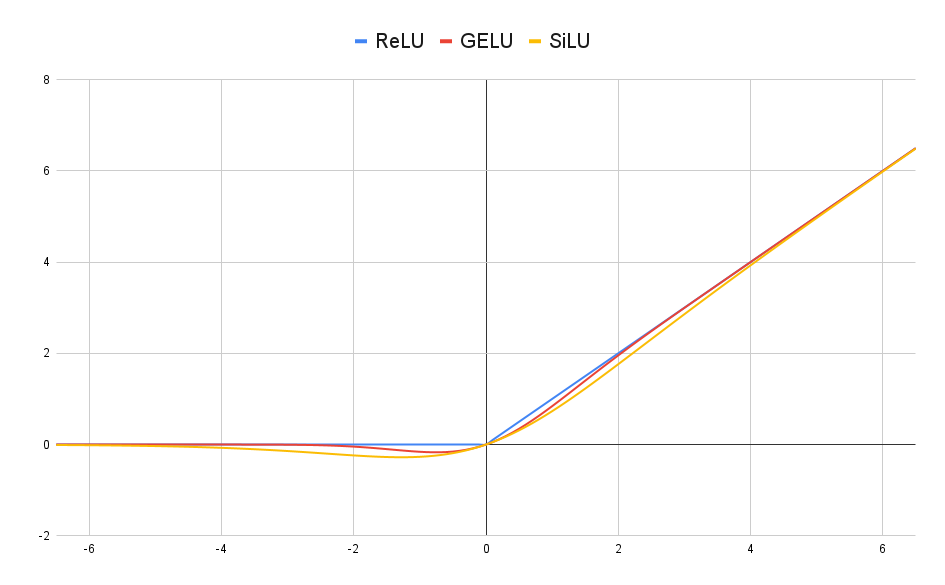}}
\caption{Use of altered sequence mixers: When the local dense sequence mixer in DMM is designed to account for the input's multi-modality, the performance remains similar or decreases. When a 1D convolution is adopted as the local sequence mixer and combined with the selective scan SSM, the performance still improves, highlighting the importance of integrating local information.}
\label{fig:activation_functions}
\end{center}
\vskip -0.2in
\end{figure}

In the Transformer architecture, the ReLU activation function is typically used following the MLP, which acts as a channel mixer. We confirmed performance improvements by changing the activation function from ReLU to GELU during experiments with the Decision Convformer on the AntMaze task. 
In the Mamba architecture, the SiLU activation function is used in the input layer preceding the selective scan SSM. Our experiments revealed that switching the activation function from SiLU to GELU in DMM did not lead to any noticeable performance gains. The reason for this can be found in Figure~\ref{fig:activation_functions}, which shows that SiLU and GELU exhibit similar functional behavior. The functional difference between ReLU and GELU is most pronounced near zero, where the majority of values lie, potentially impacting performance. Unlike ReLU, which outputs zero for negative inputs, GELU preserves some information from negative inputs. Since SiLU does not exhibit significant differences in function values compared to GELU around zero, the impact on overall model performance is likely to be minimal.

\section{Modification of the Kitchen Dataset Reward Array} \label{app:kitchen_modification}

The Franka Kitchen environment involves performing various tasks in a kitchen setting, such as moving to the position of a kettle or turning lights on and off. Each task completion earns a reward of 1 point. Since no reward is given until a task is completed, the environment provides sparse rewards.

The reward array in the Kitchen dataset represents the cumulative sum of rewards received up to each index step. For example, if a reward of 1 is received at step $n$ and another reward of 1 at step 
$m$, the reward is 0 until step $n$, 1 from step $n$ to $m-1$, and 2 from step $m$ to the final step. To convert this array into a returns-to-go (RTG) format, it is necessary to modify the array so that only $n$-th and $m$-th elements are set to 1, while all other elements are set to 0. After this modification, we computed the returns-to-go for each step to generate a new RTG array.

\begin{figure}[ht]
\vskip 0.2in
\begin{center}
\scalebox{0.4}{\includegraphics[width=\columnwidth]{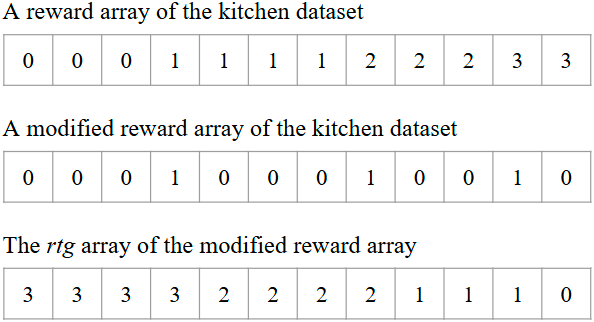}}
\caption{The arrangement of the RTG and reward array within the Kitchen dataset}
\label{fig:rtg_kitchen}
\end{center}
\vskip -0.2in
\end{figure}
\hfill

This modification aligns the input sequence with the definition of returns-to-go, which is the cumulative reward that can be obtained from the current step onward. If the rewards in the D4RL dataset's kitchen sparse dataset are used as-is, or if RTG is calculated directly from these rewards, rewards are reflected in steps where no reward was actually obtained. As the sequence length increases, this can make the total trajectory return appear inflated. To prevent this, we ensured that rewards are only represented at steps where they were actually earned and used this adjusted data to compute the RTG.

Algorithm \ref{alg:compute_rtg_kitchen} is an extension of a general RTG calculation algorithm. The 'Compute Delta' step was specifically introduced to accommodate the Kitchen dataset's RTG calculation requirements.

\begin{algorithm}[tb]
   \caption{Compute return-to-Go (RTG) for Trajectories in the Kitchen dataset}
   \label{alg:compute_rtg_kitchen}
\begin{algorithmic}
   \STATE {\bfseries Input:} Trajectories $\mathcal{T} = \{\tau_1, \tau_2, \dots, \tau_n\}$
   \STATE {\bfseries Output:} List of RTG arrays $\mathcal{R} = \{R_1, R_2, \dots, R_n\}$
   \STATE Initialize $\mathcal{R} \gets \{\}$

   \FOR{each trajectory $\tau \in \mathcal{T}$}
       \STATE Let $\mathbf{r} \gets \text{rewards in } \tau$
       \STATE \textit{\textbf{Compute deltas: $\mathbf{r} \gets \text{diff}(\mathbf{r})$ with $\mathbf{r}[0] \gets 0$}}
       \STATE Initialize $\mathbf{R} \gets \text{zeros\_like}(\mathbf{r})$
       \STATE $\mathbf{R}[-1] \gets \mathbf{r}[-1]$

       \FOR{$t = |\mathbf{r}| - 2$ {\bfseries to} $0$ (in reverse)}
           \STATE $\mathbf{R}[t] \gets \mathbf{r}[t] + \mathbf{R}[t+1]$
       \ENDFOR

       \STATE Reshape $\mathbf{R} \gets \mathbf{R}.reshape(1, -1, 1)$
       \STATE Append $\mathbf{R}$ to $\mathcal{R}$
   \ENDFOR

   \STATE {\bfseries Return:} $\mathcal{R}$
\end{algorithmic}
\end{algorithm}

\section{HiPPO initialization} \label{hippo_initialization}
Initialization with HiPPO (High-Order Polynomial Projection Operator)\cite{gu2020hippo}: HiPPO initializes the transition matrix by projecting input data onto an orthogonal polynomial basis. This design ensures that the influence of earlier input terms naturally decays due to recurrence dynamics. Unlike self-attention, which assigns weights independent of the distance between the current and past steps, HiPPO-based initialization aligns with the Markov property and distinguishes Mamba’s SSM.
Strictly speaking, Mamba does not retain the full properties of the HiPPO matrix after learning, as parameters of the transition matrix are updated during training. However, experiments indicate that HiPPO initialization yields better performance than random initialization in certain applications\cite{gu2023mamba}.

\begin{table}[t]
\renewcommand{\arraystretch}{1.1}
\caption{A performance comparison of Mamba using random and HiPPO transition matrix initialization methods}
\label{tab:hippo_init}
\vskip 0.15in
\begin{center}
\begin{small}
\begin{sc}
\resizebox{0.48\textwidth}{!}{
\begin{tabular}{lcc}
\toprule
AntMaze-UM	&Random init	&HiPPO init\\
\midrule
average	&89.3	&88.3\\
stddev	&2.1	&1.5\\
\midrule
AntMaze-UD	&Random init	&HiPPO init\\
average	&91.7	&90.7\\
stddev	&6.7	&1.2\\
\bottomrule
\end{tabular}}
\end{sc}
\end{small}
\end{center}
\vskip -0.1in
\end{table}

\begin{table}[!t]
\renewcommand{\arraystretch}{1.2}
\caption{Comparison of \textit{DMM} and \textit{Multi-modal} performance in dense reward environments (Hopper-MD, AntMaze-UM) and sparse reward environments (Kitchen-MX). Values in parentheses represent performance drops compared to \textit{DMM}.}
\label{tab:abl_mm}
\vspace{0.2cm}
\centering
\small
\begin{tabular}{l|ccc}
\toprule
& Hopper-MD & AntMaze-UM & Kitchen-MX \\
\midrule
DMM & 96.2 & 91.0 & 83.0 \\
\midrule
Multi-modal & 95.4(-0.8) & 83.0(-8.0) & 67.1(-15.9) \\
\bottomrule
\end{tabular}
\end{table}

\section{Multi-modal Token Mixers}
An offline RL dataset typically contains agent-collected data for each timestep, including states (vector), actions (vector), and rewards (scalar). Because these differ in dimension, a dense layer is used to unify them into a consistent size. So, recent state of the arts such as decision convformer~\citep{kim2024decision} often employed separate networks for each modality to improve token mixing. 

However, our experiments on the offline dataset show that multi-modality does not enhance performance in either dense or sparse reward environments. In particular, under sparse reward environments, the $rtg$ remains constant within a trajectory, rendering the $rtg$ token mixer redundant and limiting its learning utility. This redundancy leads to performance degradation in sparse environments. These findings are confirmed through the comparison of results using a multi-modal dense sequence mixer, as shown in \Cref{tab:abl_mm}.